\documentclass[letterpaper]{article}
\usepackage{aaai24}

\usepackage{natbib}
\usepackage{times}
\usepackage{helvet}
\usepackage{courier}
\usepackage[hyphens]{url}  
\usepackage{lipsum} 
\usepackage{xspace}
\usepackage{amsfonts}
\usepackage{graphicx}
\usepackage[dvipsnames]{xcolor}
\usepackage{tikz}
\usepackage{pgfplots}
\usetikzlibrary{pgfplots.statistics}
\usepackage[ruled,vlined,linesnumbered]{algorithm2e}
\usepackage{subcaption}
\usepackage{verbatim}
\usetikzlibrary{decorations.pathreplacing,calc}
\usetikzlibrary{decorations.pathreplacing,calc}
\usetikzlibrary{intersections}
\usepackage{soul}
\usepackage{pgf}
\usetikzlibrary{shapes.geometric}
\usetikzlibrary{shapes.arrows}
\usetikzlibrary{shapes,arrows,automata,positioning,fit,arrows.meta,calc}
\usetikzlibrary{decorations.markings}
\usetikzlibrary{patterns}
\usetikzlibrary{3d}
\usetikzlibrary{quotes,matrix}

\usepgfplotslibrary{fillbetween}
\pgfplotsset{compat=newest, 
	tick label style={font=\scriptsize},
	label style={font=\scriptsize},
	legend style={font=\scriptsize}}

\newcommand{\gpt}{{\sc GPT-4}\xspace}
\newcommand{\approach}{\ensuremath{\textsc{TRIP-PAL}}\xspace}

\title{\approach: Travel Planning with Guarantees by Combining Large Language Models and Automated Planners}
\author{Tomás de la Rosa, Sriram Gopalakrishnan, Alberto Pozanco, Zhen Zeng,  Daniel Borrajo\thanks{On leave from Universidad Carlos III de Madrid}}
\affiliations{J.P. Morgan AI Research}        

\begin{document}

\maketitle

\begin{abstract}
Travel planning is a complex task that involves generating a sequence of actions related to visiting places subject to constraints and maximizing some user satisfaction criteria. 
Traditional approaches rely on problem formulation in a given formal language, extracting relevant travel information from web sources, and use an adequate problem solver to generate a valid solution. As an alternative, recent Large Language Model (LLM) based approaches directly output plans from user requests using language. Although LLMs possess extensive travel domain knowledge and provide high-level information like points of interest and potential routes, current state-of-the-art models often generate plans that lack coherence, fail to satisfy constraints fully, and do not guarantee the generation of high-quality solutions.
We propose \approach, a hybrid method that combines the strengths of LLMs and automated planners, where (i) LLMs get and translate travel information and user information into data structures that can be fed into planners; and (ii) automated planners generate travel plans that guarantee constraint satisfaction and optimize for users' utility.
Our experiments across various travel scenarios show that  \approach outperforms an LLM when generating  travel plans. 
\end{abstract}

\section{Introduction}

Travel planning is a complex process that involves finding places or points of interest (POIs), incorporating real-world constraints such as visiting and travel times, as well as scheduling the POIs into a coherent trip that maximizes a given utility function.

Traditional computer-aided solutions for travel planning based on Automated Planning~\cite{HTN_temporal_travel_plan}, and Mixed-Integer Programming~\cite{MIP_travel_planner} heavily rely on manual problem formulations to represent and formalize this knowledge.
Using formal representations, different solvers can be used to generate travel plans that maximize the users' utility while satisfying the hard constraints.
Although these approaches guarantee valid travel itineraries, the formalization step can be tedious and time consuming even for domain experts.

Recent advances in LLMs such as \gpt~\cite{gpt4_tech_report} have shown promising results in tasks requiring public domain knowledge available on the internet~\cite{suri2024use_llm_knowledge_tasks,gpt4_tech_report}, where these systems are trained on.
Travel planning is one of such tasks, and LLMs seem to be good candidates as tools to generate travel itineraries due to their extensive knowledge about travel destinations, POIs and previous itineraries shared online.
However, LLMs have been shown to be bad reasoners especially in tasks that involve sequential decision making~\cite{rao_cannot_plan}; travel planning is one such type of domain. To counter this, LLMs like \gpt would have had access to many travel plans in their training data, and it can be argued that \gpt might be capable of doing a form of case-based planning(CBP)~\cite{case_based_planning_survey} for travel planning. Given a case library (set of past pairs $\langle$ problem, plan$\rangle$) and a new problem, CBP retrieves a subset of the previous plans that are most similar to the new problem, adapts them to create a new solution, that is evaluated and stored in the case library. In the context of travel planning, that could mean combining past successful travel plans in a city to make a new plan for a user's utility, and satisfy real world constraints. With the plethora of data (cases), it is feasible for an LLM to generate good travel plans; this is one of the reasons we focus on travel planning with LLMs.

In this paper we propose \approach (Travel Itinerary Planning with Planners and Language models), a hybrid method that goes beyond just using LLMs, that leverages the strengths of both LLMs and automated planners for travel planning. In a similar spirit to the work by Liu et al.~\cite{liu2023llm+} and Rao et al.~\cite{LLM_modul_plan_Rao}, our approach utilizes LLMs to parse or extract information based on user goals, such as finding POIs and time to spend at each POI. This information is then used to formulate a planning problem that captures the user's goals and general preferences. In this paper, we assume that the general preference of visiting each POI extracted by the LLM is a proxy of the user satisfaction of visiting those POIs. The formulated planning problem is then solved using an automated planner, which takes into account the real-world constraints and generates an optimal travel plan. The planner considers factors such as commute times, time to spend at a POI, and general preferences to create a plan that maximizes user satisfaction or utility while being feasible. By explicitly representing the planning problem and using an optimal planner, we ensure that the generated plans are sound (valid), comply with constraints, and are optimal. 

In this paper, we study two planning characteristics that were not covered by previous work on the relation between LLMs reasoning capabilities and planning: over-subscription and optimality. Planning under over-subscription deals with tasks where there is no possible plan that achieves all goals given as input. For instance, let us assume we give a planner 50 POIs to visit in Paris\footnote{We can easily see that this number is already a lower bound of all relevant places to visit in Paris.} in one day. There is no reasonable travel plan that can achieve all those goals under that time constraint. Therefore, a valid/optimal solution will consider the plan that achieves the subset of goals that maximize user satisfaction or utility under the given constraints.

To evaluate the effectiveness of \approach, we conduct experiments on a diverse set of travel scenarios to generate single-day travel plans.
We compare  \approach against \gpt to assess its performance in terms of plan quality and feasibility. The experiments reveal that, as the travel planning problem involves more POIs or longer travel time, \gpt tends to generate travel plans with much lower user utility than the ones generated by \approach. Specifically, \approach robustly generates valid plans that maximize user utility.

Our main contributions are as follows:
\begin{itemize}
   
    \item We investigate the use of LLMs (\gpt) for the oversubscription planning problem in the travel domain, considering real-world constraints.
    
    \item We propose a hybrid approach that combines LLMs and automated planners for the travel planning task that is an end-to-end solution with guarantees including constraint satisfaction and optimization of the utility of the travel plans. 
    
    \item We demonstrate the effectiveness of our hybrid approach through extensive experiments, showing its advantages over using only \gpt for travel plans in terms of plan utility and validity.
\end{itemize}

\section{Related Work}

\subsection{Traditional Travel Planning}
Travel recommendation and itinerary planning have been extensively studied in the literature. Numerous works have explored various algorithmic approaches to find exact or approximate optimal itineraries given points of interest (POIs) and certain constraints~\cite{berka2004designing, expertsystems08-samap, moreno2013sigtur, CitytripPlanner_ExpertSystems2011, rodriguez2013gat, ETourism_Sebastia_IJAIT09, IEEETrans_itineraryplanning, HTN_temporal_travel_plan, MIP_travel_planner}. Gavalas et al.~\cite{gavalas2014survey} provided a comprehensive survey of the research landscape in Tourist Trip Design Problems (TTDPs). They defined TTDP as the route-planning problem for tourists interested in visiting multiple POIs while accounting for constraints. The survey highlighted that TTDPs are often solved using classical optimization approaches or stochastic local search techniques. These methods aim to generate personalized and efficient travel plans catering to users' specific preferences and constraints.

To further enhance the user experience, recent works have focused on incorporating personalization techniques by harnessing crowdsourcing and social networks~\cite{DBLP:conf/aaai/ManikondaCDTK14, lucas2012hybrid, eswa-ondroad, IEEETrans_itineraryplanning, DBLP:journals/ipm/BrilhanteMNPR15, majid2014system}, as well as case-based reasoning~\cite{ricci2006case}. These methods leverage the collective knowledge and experiences of other travelers to provide more tailored and relevant itinerary recommendations. In addition, Tenemaza et al.~\cite{tenemaza2020improving} employed genetic algorithms to refine and improve proposed travel plans.

A common limitation of the aforementioned approaches is the need for manual formulation of the optimization or planning problem. More recent works use deep learning techniques for POIs and travel recommendations~\cite{halder2024survey}, lifting the requirement for manual formulation of the planning problem. However, these deep learning models do not account for optimality from a planning perspective. In contrast, our work addresses this challenge by utilizing LLMs to automatically parse user travel queries into planning problems. By adopting the method proposed by Keyder et al.~\cite{keyder2009soft} for compiling away soft goals, we can effectively solve the oversubscription problem and find the optimal travel plan without requiring manual problem formulation.

\subsection{Travel Planning with LLMs}
In recent years, a few works have explored the use of LLMs for travel planning. An early probe of LLMs such as ChatGPT on automatic travel decision-making~\cite{GPT_travel_plan_wong2023autonomous} showed LLMs' capability to improve tourists' experience in the pre-trip, en-route, and post-trip stages. However, when quantitatively benchmarked LLM agents on travel planning, it revealed that the problem is still too difficult for LLM agents, with even \gpt achieving only a low success rate~\cite{xie2024travelplanner}. To address this issue,~\citeauthor{gundawar2024robust}~(\citeyear{gundawar2024robust}) adopted the LLM-Modulo framework~\cite{kambhampati2024can} in travel planning, which iteratively improves the generated travel plan through interactions between a travel plan generation LLM and a suite of external verifiers. However, this approach does not involve formal planning or optimization tools, thus providing no guarantee of meeting all travel constraints, especially as the constraints increase~\cite{zheng2024natural}. To provide guarantees,~\citeauthor{hao2024large}~(\citeyear{hao2024large}) proposed to use LLMs to formulate the travel planning problem as a satisfiability modulo theory (SMT) problem, followed by using an SMT solver. Their methodology led to promising travel planning results where user constraints are satisfied. However, their method assumes visiting all given POIs, rather than automatically trading-off, which is the most realistic scenario, and selecting the optimal subset of POIs as in the oversubscription setting. Additionally, \approach maximizes the user's utility on top of satisfying constraints. To accommodate both constraint satisfaction and utility maximization which was not included in~\cite{xie2024travelplanner}, we created a new benchmark dataset as discussed in our experiments.

LLMs could be a valuable source of POI recommendations given their extensive knowledge, providing results similar to those acquired from public sources~\cite{DBLP:journals/ipm/BrilhanteMNPR15, majid2014system} such as Flickr and Wikipedia. However, for numeric information, such as commute time or distances between places, especially for less populated areas~\cite{roberts2023gpt4geo}, web services like Google Places API~\cite{google_places_api} are a more accurate source.

\subsection{LLMs and Planning}
When we extend the scope from travel planning to general planning, there are many more recent papers leveraging LLMs in the general planning domain. Recent works have tried using LLMs to solve planning tasks directly through prompting~\cite{valmeekam2022large,silver2024generalized}. These works conducted extensive evaluations on standard planning benchmarks and showed that LLMs perform poorly at generating valid plans. This highlights the limitations of using LLMs alone for complex planning tasks and the need for more advanced approaches.

To address this issue, researchers have explored ways to bridge LLMs and planners, such that LLMs are responsible of formulating the planning problem in standard planning languages, such as Planning Domain Definition Language (PDDL)~\cite{PDDL21-JAIR03}, while planners are in charge of solving the formulated planning problems~\cite{guan2023leveraging,liu2023llm+}. Specifically, LLMs have been shown to generate PDDL problems reasonably well~\cite{oswald2024large,pallagani2024prospects}. By leveraging the strengths of both LLMs and planners, these hybrid approaches improved the overall performance and efficiency of planning systems.

However, the aforementioned works primarily evaluated LLMs in classical planning domains such as blocksworld, which may not have had sufficient relevant data during the training stage of these LLMs. In contrast, the travel planning domain is a domain that LLMs possess extensive knowledge of, as evident in~\cite{GPT_travel_plan_wong2023autonomous}. Therefore, it is crucial to benchmark how well \gpt readily addresses the travel planning problem from both constraints and utility perspectives. To the best of our knowledge, we are the first to 1) introduce the hybrid approach of LLMs and planners to the travel planning domain and benchmark its effectiveness; 2) solve oversubscription planning tasks using such hybrid approach; and 3) benchmark the suboptimality of the travel plans generated by \gpt.

\section{Method}
\begin{figure*}
\centering
\includegraphics[width=0.78\textwidth]{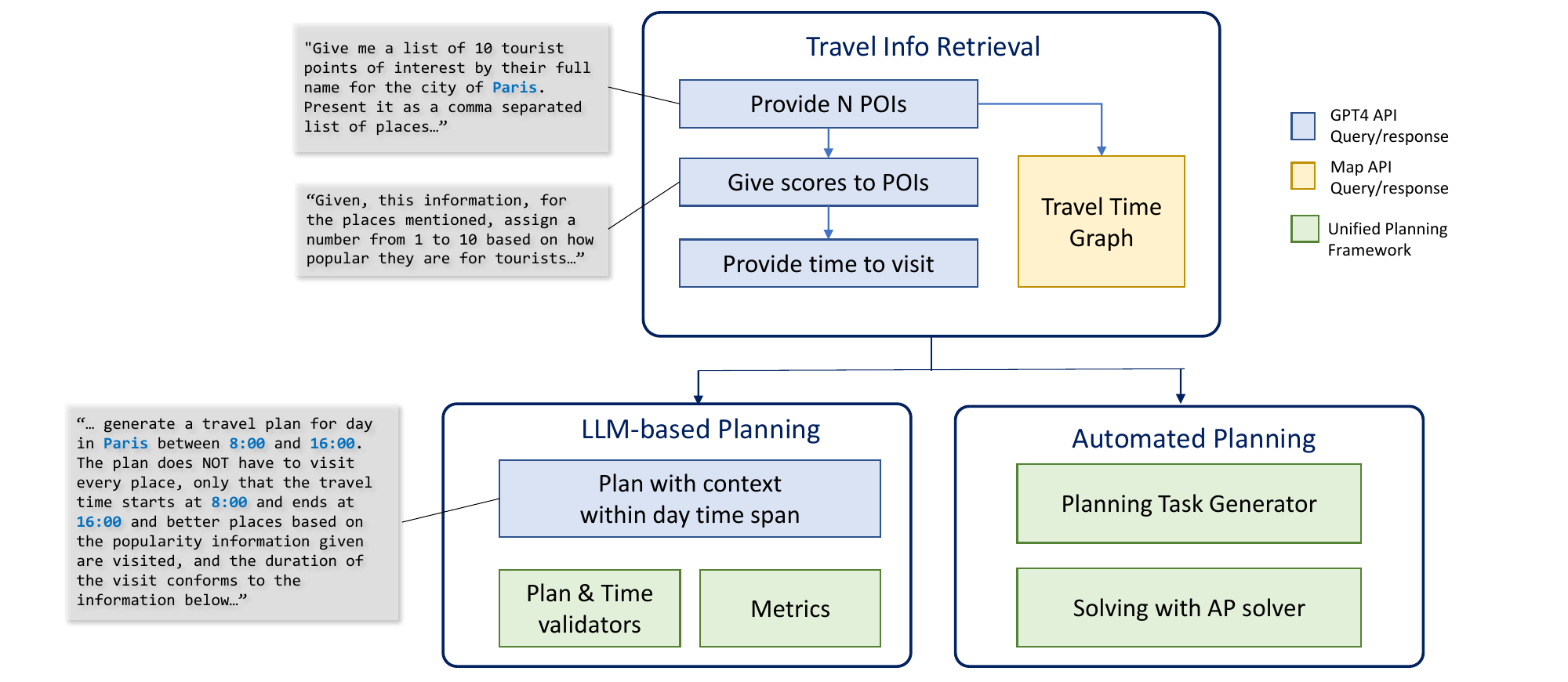}
    \caption{Components used in travel planning for \gpt and \approach.}
    \label{fig:architecture}
\end{figure*}

In this section, we describe how we extract the travel information, and how that information is used to generate plans by \gpt and \approach. The components used are illustrated in Figure~\ref{fig:architecture}.

\subsection{Retrieval of Travel Information}

The inputs to each planning episode are: the desired city, $C$; the number of POIs to consider, $N$; and the maximum number of hours the travel plan should take, $H$. The first step in our travel planning process is to extract the corresponding information to those requirements: a set of size $N$ of POIs in the city $C$ with associated relevant information. \approach uses a sequence of prompts to \gpt, that return the data it requires to compose a travel plan (excerpts from these prompts are in Figure~\ref{fig:architecture}). The result of each prompt is included in the context of the next call. The summary of steps goes as follows:
\begin{itemize}
    \item For a given \textit{city} $C$, \approach asks \gpt to return a set $P$ of $N$ tourist points of interest (POIs). Each POI $p\in P$ will become a goal of the planning problem as visit($p$).
    \item For each POI provided before, \approach gets the rating or utility of the POI by asking \gpt for the popularity expressed in a utility value between 1 and 10, with 10 as the maximum utility. It uses this value as the POI's utility, utility($p, u$). The sum of utilities of POIs in any plan will provide the total plan utility or plan value. Since we are dealing with oversubscription planning, not all POIs will necessarily be visited in the plans. 
    \item Then, \approach asks for the time one should spend in each location; this information is asked to be returned in units of 15 minutes, and will define the visit-time($p, t$) property.  
\end{itemize}

With the set of POIs returned by \gpt, \approach can query a service to get the travel time between POIs such as the google places API~\cite{google_places_api}. For this work we randomly assigned travel times between POIs to be between 15 and 60 minutes in units of 15-minutes. This gives us a travel-times graph. 
All the accumulated travel information described so far is then fed to a travel planner. The travel planning step can be a simple call to \gpt or alternatively (in \approach) a call to an automated planner. In the latter, it called after all the information is translated to a formal planning task. This flow of information is shown in Figure~\ref{fig:architecture}.

\subsection{LLM-based Tourist Plan}
All the acquired travel information (set of POIs, utility of each POI, visit time, and travel time) is concatenated  and fed as text to \gpt. \gpt is then asked to generate a travel plan for the time specified, and to optimize the travel plan based on the utilities of the points of interest passed in. We prompt \gpt to generate a day travel plan for a fixed amount of time indicated with start and end hours. To help \gpt generate valid plans in as many cases as possible, we asked it to output the travel plan using explicit time (e.g., \textit{14:00}) which empirically worked better for us, for respecting the time constraint than using AM/PM or just asking to \textit{``give a plan for x-hours''}. See an example in Figure~\ref{fig:architecture}.
For \gpt, the task of generating this plan with the previous information given in the prompt, involves selecting a subset of POIs that will fit in the schedule; this means considering the visiting time of POIs and the commute time between POIs, and then provide the best plan by utility to the user. Utility is the sum of the utilities of POIs in the plan.

\subsection{Automated Planning Task}

Automated Planning (AP) is the AI field that builds domain-independent solvers which produce sequence of actions (plans) to transform a given initial state into a  state where a set of goals have been achieved. A planning task in AP is  specified in a declarative language (typically PDDL) comprising an action model (domain) and a problem description where the initial state, the goals and the metric to optimize are defined. Oversubscription planning focuses on the class of tasks where not all goals can be achieved because the availability of bounded resources. This is the case for travel planning, since a tourist may be willing to visit more POIs that one can handle in a day itinerary. In oversubscription planning, some, or even all goals, can be deemed as \textit{soft-goals}, meaning that a plan is still valid if they are not reached in the final state. In PDDL3 this is represented with goal preferences~\cite{PDDL30}. In this case, planners  optimize for soft goals based on an utility function. Alternatively, soft goals
can be represented with classic planning tasks where high-penalty artificial 
actions are included in the domain to skip the goals that are not achieved at the end of the real plan~\cite{keyder2009soft}.  In this second case, planners optimize the cost function trying to avoid paying high penalties. 
In our setting, we set visiting all POIs as soft goals, encoding them with the latter approach.

In order to generate the planning task from the extracted information on a given travel (POIs, time constraints, driving times, ...), we used \gpt to convert the information on the POIs into dictionaries for the POI utility, travel-time and visit-time. \approach then converts these dictionaries into a PDDL planning task using the UPF planning library in python~\cite{upf}. The domain is fixed across all tasks and contains the following actions:

    \begin{itemize}
        \item Visit: the tourist spends time in a POI. For instance, visiting a park, taking a guided tour in a museum, etc.
        \item Move: A generic action representing the time spent in going from one point to another. This includes walking short distances or taking public transport.
    \end{itemize}

 Figure~\ref{fig:pddl-visit} shows the definition of the visit action in PDDL. Symbols preceded by a question mark represent variables. Preconditions define conditions that have to be true in the state in order to execute the action. Effects are literals that are expected to be true in the state after applying the action. 

\begin{figure}
\centering
\footnotesize{
\begin{verbatim}
(:action visit
    :parameters (?vloc - location 
                 ?vt0 - time 
                 ?vtvisit - time 
                 ?vtf - time)
    :precondition
      (and (normal_mode) 
           (current_time ?vt0) 
           (visited_time ?vloc ?vtvisit) 
           (user_at ?vloc) 
           (logic_sum ?vt0 ?vtvisit ?vtf))
    :effect 
      (and (visited ?vloc) 
           (not (current_time ?vt0)) 
           (current_time ?vtf) 
           (increase (total-cost) 
                     (visit_cost ?vloc))))    
\end{verbatim}
}
\caption{PDDL definition of the visit action.}
\label{fig:pddl-visit}
\end{figure}

Additionally, the model includes the artificial actions (\textit{end\_mode} and \textit{no\_visit}) needed to stop the real plan and then ignore unachieved goals~\cite{keyder2009soft} and thus, handle the oversubscription problem in the same way as in previous work~\cite{eswa-ondroad}.
The problem description incorporates as static facts the information collected in the information retrieval step regarding the time to spend at POIs and time to commute. 
Given that most current planners only optimize by minimizing a given function, the POIs' utility is reversed (\textit{MaxUtility - POI\_Utility + 1}), to translate it to cost.
Time is discretized to 15-minutes slots, so the task can be represented without numeric preconditions. Each day slot is represented as an object. Thus, domain actions include parameters for the current slot, the time slot at action termination, and delta between the two. The logic of how to assign each parameter is encoded through pre-computed static facts of the initial state.  See for instance in Figure~\ref{fig:pddl-visit} the \textit{logic\_sum} precondition, which requires the current time (\textit{vt0}) is changed to the final time slot (\textit{vtf}) when the time-to-visit slot (\textit{vtvisit}) has passed.

\begin{figure*}
    \begin{subfigure}{0.5\textwidth}
        \centering

\begin{tikzpicture}
\centering
\footnotesize
\begin{axis}[
    xlabel={\approach Utility},
    ylabel={GPT-4 Utility},
    ylabel near ticks,
    xlabel near ticks,
    legend style={xshift=-3.2cm,yshift=-0.2cm},
    enlargelimits=false,
    clip=true,
     xmin=20, xmax=70,
     ymin=20, ymax=70,
     ticklabel style ={font=\scriptsize},
     width=7cm,
     height=7cm
]

\addplot[
scatter,
    only marks,
    scatter src=explicit symbolic,
    coordinate style/.from={scale=\thisrow{mksize}},
    scatter/classes={
        popular={mark=*,Peach},
        unpopular={mark=square*, NavyBlue}
    },
] table[x=x, y=y, meta=category] {
x    y    category    city mksize
36 27 popular Tokyo_0_POI_10 0.6
38 27 popular Tokyo_1_POI_10 0.5
32 27 popular Tokyo_2_POI_10 0.5
32 21 popular Tokyo_3_POI_10 0.7
36 27 popular Tokyo_4_POI_10 0.6
39 21 popular Paris_0_POI_10 1.2
39 21 popular Paris_1_POI_10 1.2
37 21 popular Paris_2_POI_10 0.5
39 21 popular Paris_3_POI_10 1.2
39 21 popular Paris_4_POI_10 1.2
39 21 popular Barcelona_0_POI_10 1.2
42 36 popular Barcelona_1_POI_10 0.6
42 21 popular Barcelona_2_POI_10 0.7
42 36 popular Barcelona_3_POI_10 0.6
45 21 popular Barcelona_4_POI_10 0.6
43 21 popular New_York_City_0_POI_10 0.9999999999999999
41 21 popular New_York_City_1_POI_10 0.8999999999999999
42 21 popular New_York_City_2_POI_10 0.7
43 21 popular New_York_City_3_POI_10 0.9999999999999999
41 21 popular New_York_City_4_POI_10 0.8999999999999999
43 21 popular London_0_POI_10 0.9999999999999999
44 21 popular London_1_POI_10 0.7999999999999999
45 21 popular London_2_POI_10 0.6
44 21 popular London_3_POI_10 0.7999999999999999
44 21 popular London_4_POI_10 0.7999999999999999
29 21 popular Cape_Town_0_POI_10 0.8999999999999999
32 21 popular Cape_Town_1_POI_10 0.7
29 21 popular Cape_Town_2_POI_10 0.8999999999999999
32 21 popular Cape_Town_3_POI_10 0.7
29 21 popular Cape_Town_4_POI_10 0.8999999999999999
36 21 popular Amsterdam_0_POI_10 0.9999999999999999
24 21 popular Amsterdam_1_POI_10 0.6
38 21 popular Amsterdam_2_POI_10 0.6
24 21 popular Amsterdam_3_POI_10 0.6
38 21 popular Amsterdam_4_POI_10 0.6
34 21 popular Toronto_0_POI_10 0.7999999999999999
34 21 popular Toronto_1_POI_10 0.7999999999999999
34 21 popular Toronto_2_POI_10 0.7999999999999999
36 21 popular Toronto_3_POI_10 0.9999999999999999
36 21 popular Toronto_4_POI_10 0.9999999999999999
53 21 popular Rome_0_POI_10 0.7
53 21 popular Rome_1_POI_10 0.7
62 21 popular Rome_2_POI_10 0.6
53 21 popular Rome_3_POI_10 0.7
62 21 popular Rome_4_POI_10 0.6
59 21 popular Berlin_0_POI_10 0.8999999999999999
59 21 popular Berlin_1_POI_10 0.8999999999999999
59 21 popular Berlin_2_POI_10 0.8999999999999999
59 21 popular Berlin_3_POI_10 0.8999999999999999
59 21 popular Berlin_4_POI_10 0.8999999999999999
51 21 unpopular Cienfuegos,_Cuba_0_POI_10 0.7999999999999999
51 21 unpopular Cienfuegos,_Cuba_1_POI_10 0.7999999999999999
51 21 unpopular Cienfuegos,_Cuba_2_POI_10 0.7999999999999999
48 37 unpopular Cienfuegos,_Cuba_3_POI_10 0.5
51 21 unpopular Cienfuegos,_Cuba_4_POI_10 0.7999999999999999
33 21 unpopular Yogyakarta,_Indonesia_0_POI_10 0.5
34 21 unpopular Yogyakarta,_Indonesia_1_POI_10 0.7999999999999999
29 21 unpopular Yogyakarta,_Indonesia_2_POI_10 0.8999999999999999
30 21 unpopular Yogyakarta,_Indonesia_3_POI_10 0.5
29 21 unpopular Yogyakarta,_Indonesia_4_POI_10 0.8999999999999999
47 21 unpopular Matera,_Italy_0_POI_10 0.9999999999999999
47 21 unpopular Matera,_Italy_1_POI_10 0.9999999999999999
47 21 unpopular Matera,_Italy_2_POI_10 0.9999999999999999
47 21 unpopular Matera,_Italy_3_POI_10 0.9999999999999999
47 21 unpopular Matera,_Italy_4_POI_10 0.9999999999999999
43 21 unpopular Luang_Prabang,_Laos_0_POI_10 0.9999999999999999
42 21 unpopular Luang_Prabang,_Laos_1_POI_10 0.7
49 21 unpopular Luang_Prabang,_Laos_2_POI_10 0.6
48 21 unpopular Luang_Prabang,_Laos_3_POI_10 0.7
47 21 unpopular Luang_Prabang,_Laos_4_POI_10 0.9999999999999999
36 21 unpopular Salzburg,_Austria_0_POI_10 0.9999999999999999
36 21 unpopular Salzburg,_Austria_1_POI_10 0.9999999999999999
42 37 unpopular Salzburg,_Austria_2_POI_10 0.5
36 21 unpopular Salzburg,_Austria_3_POI_10 0.9999999999999999
41 21 unpopular Salzburg,_Austria_4_POI_10 0.8999999999999999
46 21 unpopular Valparaíso,_Chile_0_POI_10 0.6
46 35 unpopular Valparaíso,_Chile_1_POI_10 0.5
46 21 unpopular Valparaíso,_Chile_2_POI_10 0.6
44 21 unpopular Valparaíso,_Chile_3_POI_10 0.7999999999999999
46 42 unpopular Valparaíso,_Chile_4_POI_10 0.5
51 51 unpopular Zadar,_Croatia_0_POI_10 0.5
49 21 unpopular Zadar,_Croatia_1_POI_10 0.6
55 51 unpopular Zadar,_Croatia_2_POI_10 0.5
48 21 unpopular Zadar,_Croatia_3_POI_10 0.7
48 21 unpopular Zadar,_Croatia_4_POI_10 0.7
39 21 unpopular Bergen,_Norway_0_POI_10 1.2
39 21 unpopular Bergen,_Norway_1_POI_10 1.2
39 21 unpopular Bergen,_Norway_2_POI_10 1.2
41 21 unpopular Bergen,_Norway_3_POI_10 0.8999999999999999
40 35 unpopular Bergen,_Norway_4_POI_10 0.5
43 21 unpopular Hoi_An,_Vietnam_0_POI_10 0.9999999999999999
43 21 unpopular Hoi_An,_Vietnam_1_POI_10 0.9999999999999999
41 21 unpopular Hoi_An,_Vietnam_2_POI_10 0.8999999999999999
40 21 unpopular Hoi_An,_Vietnam_3_POI_10 0.6
40 21 unpopular Hoi_An,_Vietnam_4_POI_10 0.6
57 21 unpopular Colonia_del_Sacramento,_Uruguay_0_POI_10 0.5
60 21 unpopular Colonia_del_Sacramento,_Uruguay_1_POI_10 0.5
57 51 unpopular Colonia_del_Sacramento,_Uruguay_2_POI_10 0.5
66 21 unpopular Colonia_del_Sacramento,_Uruguay_3_POI_10 0.5
55 21 unpopular Colonia_del_Sacramento,_Uruguay_4_POI_10 0.5
};


		\draw[color=black] (axis cs:0,0) -- (axis cs:70,70);
		\draw[color=black, dashed] (axis cs:0,8.75) -- (axis cs:61.25,70);
        \draw[color=black, loosely dashed] (axis cs:0,17.5) -- (axis cs:52.5,70);
        \draw[color=black, dashed] (axis cs:8.75,0) -- (axis cs:70,61.25);
        \draw[color=black, loosely dashed] (axis cs:17.5,0) -- (axis cs:70,52.5);
        \draw[color=red] (axis cs:21,21) -- (axis cs:70,21);

\legend{popular, unpopular}

\end{axis}
\end{tikzpicture}
        \caption{With validity check.}
        \label{fig:with_validity_check}
    \end{subfigure}\hfill
    \begin{subfigure}{0.5\textwidth}
        \centering

\begin{tikzpicture}
\centering
\footnotesize
\begin{axis}[
    xlabel={\approach Utility},
    ylabel={GPT-4 Utility},
    ylabel near ticks,
    xlabel near ticks,
    legend style={xshift=-3.2cm,yshift=-0.2cm},
    enlargelimits=false,
    clip=true,
     xmin=20, xmax=70,
     ymin=20, ymax=70,
     ticklabel style ={font=\scriptsize},
     width=7cm,
     height=7cm
]

\addplot[
scatter,
    only marks,
    scatter src=explicit symbolic,
    coordinate style/.from={scale=\thisrow{mksize}},
    scatter/classes={
        popular={mark=*,Peach},
        unpopular={mark=square*, NavyBlue}
    },
] table[x=x, y=y, meta=category] {
x    y    category    city mksize
36 27 popular Tokyo_0_POI_10 0.6
38 27 popular Tokyo_1_POI_10 0.5
32 27 popular Tokyo_2_POI_10 0.6
32 27 popular Tokyo_3_POI_10 0.6
36 27 popular Tokyo_4_POI_10 0.6
39 35 popular Paris_0_POI_10 1.0999999999999999
39 35 popular Paris_1_POI_10 1.0999999999999999
37 35 popular Paris_2_POI_10 0.5
39 35 popular Paris_3_POI_10 1.0999999999999999
39 35 popular Paris_4_POI_10 1.0999999999999999
39 42 popular Barcelona_0_POI_10 0.5
42 36 popular Barcelona_1_POI_10 0.6
42 42 popular Barcelona_2_POI_10 0.5
42 36 popular Barcelona_3_POI_10 0.6
45 42 popular Barcelona_4_POI_10 0.5
43 39 popular New_York_City_0_POI_10 0.6
41 38 popular New_York_City_1_POI_10 0.6
42 38 popular New_York_City_2_POI_10 0.5
43 39 popular New_York_City_3_POI_10 0.6
41 38 popular New_York_City_4_POI_10 0.6
43 45 popular London_0_POI_10 0.5
44 38 popular London_1_POI_10 0.7
45 38 popular London_2_POI_10 0.5
44 38 popular London_3_POI_10 0.7
44 38 popular London_4_POI_10 0.7
29 29 popular Cape_Town_0_POI_10 0.7
32 29 popular Cape_Town_1_POI_10 0.6
29 29 popular Cape_Town_2_POI_10 0.7
32 29 popular Cape_Town_3_POI_10 0.6
29 29 popular Cape_Town_4_POI_10 0.7
36 35 popular Amsterdam_0_POI_10 0.5
24 27 popular Amsterdam_1_POI_10 0.6
38 35 popular Amsterdam_2_POI_10 0.6
24 27 popular Amsterdam_3_POI_10 0.6
38 35 popular Amsterdam_4_POI_10 0.6
34 29 popular Toronto_0_POI_10 0.7
34 29 popular Toronto_1_POI_10 0.7
34 29 popular Toronto_2_POI_10 0.7
36 38 popular Toronto_3_POI_10 0.5
36 29 popular Toronto_4_POI_10 0.5
53 40 popular Rome_0_POI_10 0.7
53 40 popular Rome_1_POI_10 0.7
62 40 popular Rome_2_POI_10 0.6
53 40 popular Rome_3_POI_10 0.7
62 40 popular Rome_4_POI_10 0.6
59 54 popular Berlin_0_POI_10 0.8999999999999999
59 54 popular Berlin_1_POI_10 0.8999999999999999
59 54 popular Berlin_2_POI_10 0.8999999999999999
59 54 popular Berlin_3_POI_10 0.8999999999999999
59 54 popular Berlin_4_POI_10 0.8999999999999999
51 43 unpopular Cienfuegos,_Cuba_0_POI_10 0.7999999999999999
51 43 unpopular Cienfuegos,_Cuba_1_POI_10 0.7999999999999999
51 43 unpopular Cienfuegos,_Cuba_2_POI_10 0.7999999999999999
48 37 unpopular Cienfuegos,_Cuba_3_POI_10 0.5
51 43 unpopular Cienfuegos,_Cuba_4_POI_10 0.7999999999999999
33 37 unpopular Yogyakarta,_Indonesia_0_POI_10 0.5
34 37 unpopular Yogyakarta,_Indonesia_1_POI_10 0.5
29 28 unpopular Yogyakarta,_Indonesia_2_POI_10 0.6
30 28 unpopular Yogyakarta,_Indonesia_3_POI_10 0.5
29 28 unpopular Yogyakarta,_Indonesia_4_POI_10 0.6
47 34 unpopular Matera,_Italy_0_POI_10 0.8999999999999999
47 34 unpopular Matera,_Italy_1_POI_10 0.8999999999999999
47 34 unpopular Matera,_Italy_2_POI_10 0.8999999999999999
47 34 unpopular Matera,_Italy_3_POI_10 0.8999999999999999
47 34 unpopular Matera,_Italy_4_POI_10 0.8999999999999999
43 46 unpopular Luang_Prabang,_Laos_0_POI_10 0.5
42 46 unpopular Luang_Prabang,_Laos_1_POI_10 0.5
49 46 unpopular Luang_Prabang,_Laos_2_POI_10 0.5
48 45 unpopular Luang_Prabang,_Laos_3_POI_10 0.5
47 45 unpopular Luang_Prabang,_Laos_4_POI_10 0.5
36 39 unpopular Salzburg,_Austria_0_POI_10 0.7
36 39 unpopular Salzburg,_Austria_1_POI_10 0.7
42 37 unpopular Salzburg,_Austria_2_POI_10 0.5
36 39 unpopular Salzburg,_Austria_3_POI_10 0.7
41 45 unpopular Salzburg,_Austria_4_POI_10 0.5
46 42 unpopular Valparaíso,_Chile_0_POI_10 0.7
46 35 unpopular Valparaíso,_Chile_1_POI_10 0.5
46 42 unpopular Valparaíso,_Chile_2_POI_10 0.7
44 33 unpopular Valparaíso,_Chile_3_POI_10 0.5
46 42 unpopular Valparaíso,_Chile_4_POI_10 0.7
51 51 unpopular Zadar,_Croatia_0_POI_10 0.5
49 51 unpopular Zadar,_Croatia_1_POI_10 0.5
55 51 unpopular Zadar,_Croatia_2_POI_10 0.5
48 49 unpopular Zadar,_Croatia_3_POI_10 0.6
48 49 unpopular Zadar,_Croatia_4_POI_10 0.6
39 35 unpopular Bergen,_Norway_0_POI_10 1.0999999999999999
39 35 unpopular Bergen,_Norway_1_POI_10 1.0999999999999999
39 35 unpopular Bergen,_Norway_2_POI_10 1.0999999999999999
41 35 unpopular Bergen,_Norway_3_POI_10 0.5
40 35 unpopular Bergen,_Norway_4_POI_10 0.5
43 36 unpopular Hoi_An,_Vietnam_0_POI_10 0.6
43 36 unpopular Hoi_An,_Vietnam_1_POI_10 0.6
41 36 unpopular Hoi_An,_Vietnam_2_POI_10 0.5
40 36 unpopular Hoi_An,_Vietnam_3_POI_10 0.5
40 37 unpopular Hoi_An,_Vietnam_4_POI_10 0.5
57 51 unpopular Colonia_del_Sacramento,_Uruguay_0_POI_10 0.6
60 54 unpopular Colonia_del_Sacramento,_Uruguay_1_POI_10 0.5
57 51 unpopular Colonia_del_Sacramento,_Uruguay_2_POI_10 0.6
66 64 unpopular Colonia_del_Sacramento,_Uruguay_3_POI_10 0.5
55 49 unpopular Colonia_del_Sacramento,_Uruguay_4_POI_10 0.5
};


		\draw[color=black] (axis cs:0,0) -- (axis cs:70,70);
		\draw[color=black, dashed] (axis cs:0,8.75) -- (axis cs:61.25,70);
        \draw[color=black, loosely dashed] (axis cs:0,17.5) -- (axis cs:52.5,70);
        \draw[color=black, dashed] (axis cs:8.75,0) -- (axis cs:70,61.25);
        \draw[color=black, loosely dashed] (axis cs:17.5,0) -- (axis cs:70,52.5);

\legend{popular, unpopular}

\end{axis}
\end{tikzpicture}
        \caption{Without validity check.}
        \label{fig:without_validity_check}
    \end{subfigure}
    \caption{Utility of the plans returned by \approach and \gpt. Points below the solid diagonal indicate that \approach generated a higher quality plan than \gpt for the given travel planning task. Points at the horizontal red line indicate that \gpt generated an invalid plan.}
    \label{fig:standard_day}
\end{figure*}
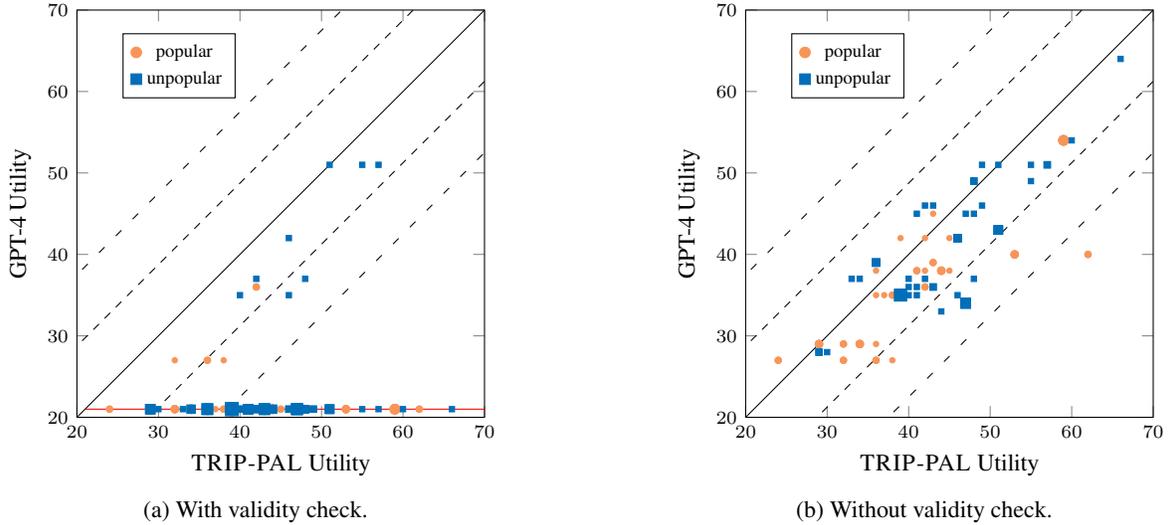

\section{Evaluation}

\subsection{Experimental Setting}

\paragraph{Approaches.}
We compare \approach against an LLM-based baseline, \gpt.
\approach uses Fast Downward~\cite{helmert06} as the AI planner, as provided in the Unified Planning Framework~\cite{upf}.
We use the \textsc{seq-opt-lmcut} configuration of Fast Downward, which runs $\mbox{A}^*$ with the admissible \textsc{lmcut} heuristic to compute plans that are guaranteed to be valid and optimal.
Experiments were run on a EC2 T3-medium instance with 4Gb of RAM.

\paragraph{Benchmark.}
We sampled $20$ cities for our experiments by asking \gpt to generate $10$ popular destinations and $10$ less popular ones.
This selection ensures a comprehensive comparison across various travel contexts.
For each city, we considered $N$ POIs located within or near the city limits. 
These POIs serve as potential destinations for the travel plans.
We ask both approaches, \gpt and \approach for one day travel plans involving at most $H$ tourism hours.

\paragraph{Metrics.}
We evaluate the generated travel plans using a set of metrics that assess their feasibility and quality. These metrics include:
\begin{itemize}
    
    \item \textit{Plan Validity}. \approach plans will be valid by definition, since the planner we use is sound. However, this is not the case for the plans generated by \gpt. We check whether LLM-based plans are valid or not in terms of action applicability and time constraints. For the applicability check, we translate each \gpt plan into PDDL output format and use the UPF validator~\cite{upf} that returns true if the plan is valid, or false otherwise.
    For the time constraints, we developed a validation function that checks if (1) the plan fits within $H$ tourism hours, and (2) each action lasts at least the time collected in the retrieval phase, i.e., travel and visit times are respected. This check is performed separately from the validator call because we verify that the time spend in each POI is at least the time required, and not the exact time. This is not directly achievable with the object representation we use for time slots. 
    If any of these constraints is violated, the \gpt-based plan is considered invalid.  
    \item \textit{Plan Utility}. We define the utility of a travel plan as the sum of the utility of the POIs it visits.

    \item \textit{Runtime}. Time in seconds needed to generate a travel plan. 
    This time comprises all the steps in each approach's pipeline.
\end{itemize}

\subsection{Results}

\begin{figure*}[thb]
    \begin{subfigure}{0.25\textwidth}
    \centering
    \begin{tikzpicture}
\scriptsize
\begin{axis}[
    ylabel={\#POIs}, 
    xtick={1,2},
    xticklabels={\gpt,\approach}, 
    xmin=0, xmax=3, 
    ymin=0, ymax=10, 
    height=5cm, 
    width=5cm,
    boxplot/draw direction=y, 
    xlabel near ticks,
    ylabel near ticks
]
\addplot+[boxplot,solid,color=NavyBlue,mark options={color=NavyBlue}] coordinates {
(1,3)
(1,3)
(1,3)
(1,3)
(1,4)
(1,4)
(1,4)
(1,4)
(1,4)
(1,5)
(1,6)
(1,6)
(1,4)
(1,7) };

\addplot+[boxplot,solid,color=ForestGreen,mark options={color=ForestGreen}] coordinates {
(2,5)
(2,5)
(2,4)
(2,5)
(2,5)
(2,5)
(2,6)
(2,5)
(2,6)
(2,6)
(2,6)
(2,7)
(2,5)
(2,8)};


\node [NavyBlue] at (1.00,4.28) {*};
\node [ForestGreen] at (2,5.57) {*};

\end{axis}
\end{tikzpicture}
    \caption{Number of POIs visited.}
    \label{fig:pois_number}
    \end{subfigure}\hfill
    \begin{subfigure}{0.25\textwidth}
    \centering
    \begin{tikzpicture}
\scriptsize
\begin{axis}[
    ylabel={Avg. Utility}, 
    xtick={1,2},
    xticklabels={\gpt,\approach}, 
    xmin=0, xmax=3, 
    ymin=0, ymax=12, 
    height=5cm, 
    width=5cm,
    boxplot/draw direction=y, 
    xlabel near ticks,
    ylabel near ticks
]
\addplot+[boxplot,solid,color=NavyBlue,mark options={color=NavyBlue}] coordinates {
(1,9.0)
(1,9.0)
(1,9.0)
(1,9.0)
(1,9.0)
(1,9.0)
(1,9.25)
(1,9.25)
(1,8.75)
(1,8.4)
(1,8.5)
(1,8.5)
(1,8.75)
(1,7.285714285714286) };

\addplot+[boxplot,solid,color=ForestGreen,mark options={color=ForestGreen}] coordinates {
(2,7.2)
(2,7.6)
(2,8.0)
(2,7.2)
(2,8.4)
(2,8.4)
(2,8.0)
(2,8.4)
(2,7.666666666666667)
(2,7.666666666666667)
(2,8.5)
(2,7.857142857142857)
(2,8.0)
(2,7.125)};

\node [NavyBlue] at (1,8.76) {*};
\node [ForestGreen] at (2,7.85) {*};

\end{axis}
\end{tikzpicture}
    \caption{Average POI utility.}
    \label{fig:avg_utility}
    \end{subfigure}\hfill
    \begin{subfigure}{0.25\textwidth}
    \centering
    \begin{tikzpicture}
\scriptsize
\begin{axis}[
    ylabel={Max. Utility}, 
    xtick={1,2},
    xticklabels={\gpt,\approach}, 
    xmin=0, xmax=3, 
    ymin=0, ymax=12, 
    height=5cm, 
    width=5cm,
    boxplot/draw direction=y, 
    xlabel near ticks,
    ylabel near ticks
]
\addplot+[boxplot,solid,color=NavyBlue,mark options={color=NavyBlue}] coordinates {
(1,10)
(1,10)
(1,10)
(1,10)
(1,10)
(1,10)
(1,10)
(1,10)
(1,10)
(1,10)
(1,10)
(1,10)
(1,10)
(1,9) };

\addplot+[boxplot,solid,color=ForestGreen,mark options={color=ForestGreen}] coordinates {
(2,8)
(2,9)
(2,9)
(2,8)
(2,10)
(2,10)
(2,10)
(2,10)
(2,10)
(2,10)
(2,10)
(2,10)
(2,10)
(2,9)};


\node [NavyBlue] at (1,9.92) {*};
\node [ForestGreen] at (2,9.50) {*};

\end{axis}
\end{tikzpicture}
    \caption{Maximum POI utility.}
    \label{fig:max_utility}
    \end{subfigure}\hfill
    \begin{subfigure}{0.25\textwidth}
    \centering
    \begin{tikzpicture}
\scriptsize
\begin{axis}[
    ylabel={Min. Utility}, 
    xtick={1,2},
    xticklabels={\gpt,\approach}, 
    xmin=0, xmax=3, 
    ymin=0, ymax=12, 
    height=5cm, 
    width=5cm,
    boxplot/draw direction=y, 
    xlabel near ticks,
    ylabel near ticks
]
\addplot+[boxplot,solid,color=NavyBlue,mark options={color=NavyBlue}] coordinates {
(1,8)
(1,8)
(1,8)
(1,8)
(1,8)
(1,8)
(1,8)
(1,8)
(1,8)
(1,7)
(1,7)
(1,7)
(1,8)
(1,5) };

\addplot+[boxplot,solid,color=ForestGreen,mark options={color=ForestGreen}] coordinates {
(2,6)
(2,6)
(2,7)
(2,6)
(2,7)
(2,7)
(2,6)
(2,6)
(2,6)
(2,6)
(2,7)
(2,6)
(2,6)
(2,5)};

\node [NavyBlue] at (1,7.57) {*};
\node [ForestGreen] at (2,6.21) {*};

\end{axis}
\end{tikzpicture}
    \caption{Minimum POI utility.}
    \label{fig:min_utility}
    \end{subfigure}\hfill
    \caption{Distribution of the number of POIs visited by each plan, as well as their average, maximum, and minimum utility.}
    \label{fig:all}
\end{figure*}
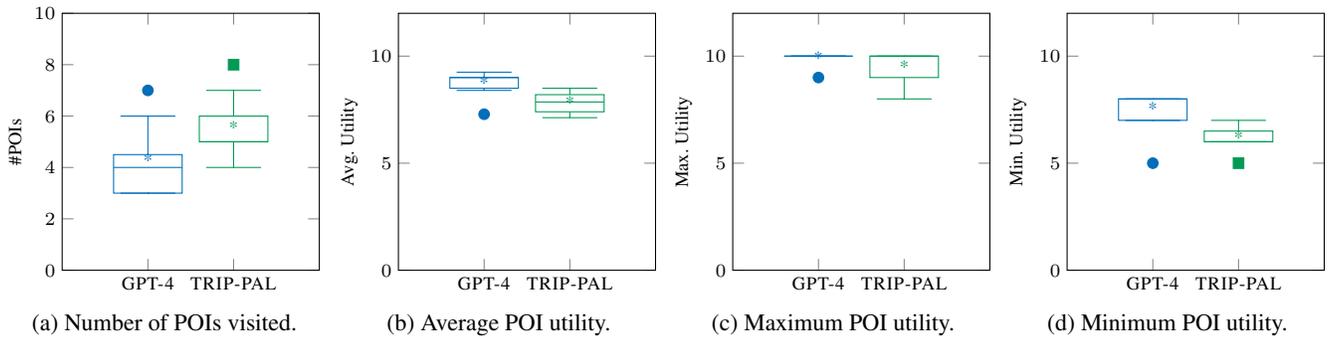

\paragraph{Standard Day Travel Planning.}
Our first experiment aims to analyze the performance of both approaches when doing \emph{standard} day travel planning.
We define this as deciding the POIs to visit (and in which order) in $H=8$ hours of tourism (leaving meals aside) among a set of $N=10$ candidate POIs.
We believe this is a standard travel planning setting faced by people when preparing for visiting a city.

We generate $5$ different problems --each problem is a set of POIs with their associated utilities and visiting times, and travel times between POIs-- for each of the $20$ cities comprising our  benchmark, thus having $100$ travel planning tasks where we can compare \approach and \gpt.
Figure~\ref{fig:with_validity_check} shows this comparison by plotting the utility of the plan computed by \approach ($x$ axis) and \gpt ($y$ axis) for each task.
Points below the solid diagonal indicate that \approach generated a higher quality plan than \gpt for the given travel planning task.
When \gpt generates an invalid plan, we associate it a utility value lower than the utility obtained by \approach in any task. We used a value of 21 so that it can be easily visualized.
These are the points falling at the horizontal red line.

As we can see, \approach generated higher utility plans than \gpt for all the tasks.
\gpt returned $14$ valid plans out of the $100$ tasks, clearly indicating that it struggles to generate travel plans that satisfy hard constraints.

Taking a closer look to the sources of invalidity, they are distributed as follows.
Across the $86$ invalid plans, \gpt did not respect a visiting time in $81$, a traveling time in $27$, and the maximum number of hours devoted to tourism in $12$ of the plans.
In $25$ out of the $86$ cases, \gpt generated plans with more than one invalidity source, with a maximum of $6$ invalid actions in some itineraries.
The suggested time to visit POIs and travel between them in these invalid plans was on average $0.48\pm0.22$ of the time required to visit/travel, with extreme cases where this time deviation went as low as $0.08$.
This was the case of a \gpt plan that suggested to spend just $15$ minutes to visit the Toronto Islands, while it was asked to allow at least $3$ hours.

This is a blatant case where \gpt plan is clearly not executable in practice.
However, in some cases the plan is invalid due to small visiting/traveling time differences.
For example, one \gpt trip is invalid due to visiting the Tower of London for $1.5$ hours rather than $2$. 
Even if incorrect from a hard constraints point of view, that plan would arguably be realistic and executable in practice, so we decided to treat \gpt favourably by removing the hard constraint checks, and report the utility of the resulting plans.
These results are shown in Figure~\ref{fig:without_validity_check}.
In this setting, \approach is still generating plans with higher utility than \gpt in $79$ of the $100$ problems.
This highlights that, even when we do not constrain \gpt to follow some guidelines, it is still generating worse travel itineraries than \approach, whose plans are guaranteed to be sound and optimal.

Focusing back on the $14$ tasks for which \gpt generated a valid travel plan, \approach's plans have on average of $1.19\pm0.12$ times more utility.

Figure~\ref{fig:all} shows the distribution of the number of POIs visited by the plans suggested by both approaches, as well as their average, maximum and minimum utility.
As we can see, while \gpt suggests visiting an average of $4.2$ POIs, \approach suggests $5.5$, with trips in which it manages to fit $8$ POIs in small cities such as Colonia del Sacramento, Uruguay.
By inspection, the \gpt solutions tend to involve visiting $3$ to $5$ of the highest utility places, which allows it to obtain a higher average utility of the visited POIs in a plan (see Figure~\ref{fig:avg_utility}).
On the other hand, \approach also considers POIs with slightly lower utility (see Figure~\ref{fig:min_utility}) if they can be visited within the given tourism hours, resulting into optimal travel plans that maximize utility.

\paragraph{Scalability.}

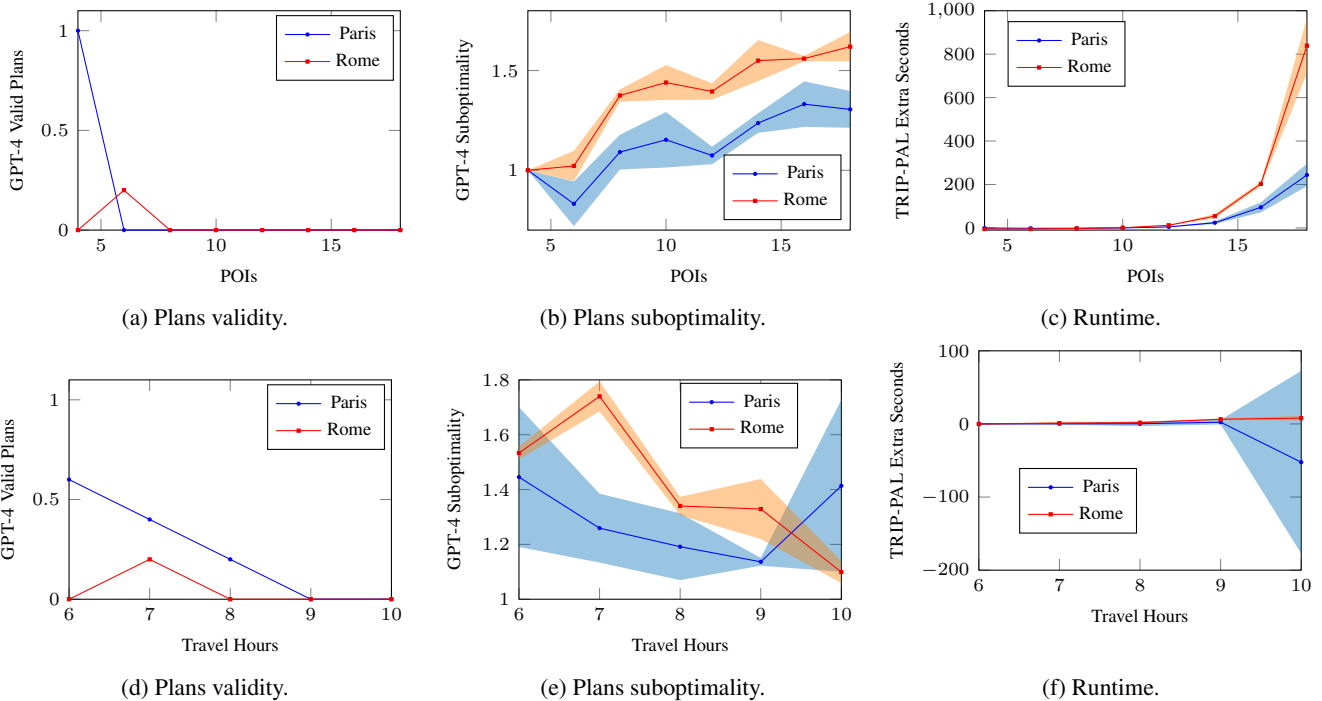
\begin{figure*}
    \begin{subfigure}{0.33\textwidth}
    \centering
    \begin{tikzpicture}
	\begin{axis}[
	width=\textwidth,
	height=4.5cm,
	xmin=4,
	xmax=18,
	ymin=0,
	ymax=1.1,
	%
	%
	xlabel=POIs,
	ylabel=\gpt Valid Plans,
	%
	mark options={scale=0.3},
	%
	%
	%
	]
	\addplot
	table [x=pois,y=Paris,col sep=comma] 
	{tikz_figures/data/varying_pois_invalidity.csv};
	\addplot
	table [x=pois,y=Rome,col sep=comma] 
	{tikz_figures/data/varying_pois_invalidity.csv};
        \legend{Paris, Rome}
	\end{axis}
	\end{tikzpicture}
    \caption{Plans validity.}
    \label{fig:scalability_POI_validity}
    \end{subfigure}\hfill
    \begin{subfigure}{0.33\textwidth}
    \centering
    \begin{tikzpicture}
	\begin{axis}[
	width=\textwidth,
	height=4.5cm,
	xmin=4,
	xmax=18,
	ymin=0.7,
	ymax=1.8,
 legend style={at={(axis cs:12.5,0.75)},anchor=south west},
	%
	%
	xlabel=POIs,
	ylabel=\gpt Suboptimality,
	%
	mark options={scale=0.3},
	%
	%
	%
	]
	\addplot
	table [x=pois,y=Paris_avg,col sep=comma] 
	{tikz_figures/data/varying_pois_suboptimality.csv};
	\addplot[name path=top,draw=none,forget plot]
	table [name path=top,x=pois,y expr=\thisrow{Paris_avg}+\thisrow{Paris_std},col sep=comma] 
	{tikz_figures/data/varying_pois_suboptimality.csv};
	\addplot[name path=bot,draw=none,forget plot]
	table [name path=top,x=pois,y expr=\thisrow{Paris_avg}-\thisrow{Paris_std},col sep=comma] 
	{tikz_figures/data/varying_pois_suboptimality.csv};
	\addplot[forget plot, draw=none,opacity=0.4,pattern=north east lines,fill=NavyBlue]
	fill between[of=top and bot];
	\addplot
	table [x=pois,y=Rome_avg,col sep=comma] 
	{tikz_figures/data/varying_pois_suboptimality.csv};
	\addplot[name path=top,draw=none,forget plot]
	table [name path=top,x=pois,y expr=\thisrow{Rome_avg}+\thisrow{Rome_std},col sep=comma] 
	{tikz_figures/data/varying_pois_suboptimality.csv};
	\addplot[name path=bot,draw=none,forget plot]
	table [name path=top,x=pois,y expr=\thisrow{Rome_avg}-\thisrow{Rome_std},col sep=comma] 
	{tikz_figures/data/varying_pois_suboptimality.csv};
	\addplot [forget plot, draw=none,opacity=0.4,pattern=north east lines,fill=orange]
	fill between[of=top and bot];
        \legend{Paris, Rome}
	\end{axis}
	\end{tikzpicture}
    \caption{Plans suboptimality.}
    \label{fig:scalability_POI_suboptimality}
    \end{subfigure}\hfill
    \begin{subfigure}{0.33\textwidth}
    \centering
    \begin{tikzpicture}
	\begin{axis}[
	width=\textwidth,
	height=4.5cm,
	xmin=4,
	xmax=18,
	ymin=-10,
	ymax=1000,
         legend style={at={(axis cs:5,650)},anchor=south west},
	%
	%
	xlabel=POIs,
	ylabel=\approach Extra Seconds,
	%
	mark options={scale=0.3},
	%
	%
	%
	]
	\addplot
	table [x=pois,y=Paris_avg,col sep=comma] 
	{tikz_figures/data/varying_pois_runtime_overhead.csv};
	\addplot[name path=top,draw=none,forget plot]
	table [name path=top,x=pois,y expr=\thisrow{Paris_avg}+\thisrow{Paris_std},col sep=comma] 
	{tikz_figures/data/varying_pois_runtime_overhead.csv};
	\addplot[name path=bot,draw=none,forget plot]
	table [name path=top,x=pois,y expr=\thisrow{Paris_avg}-\thisrow{Paris_std},col sep=comma] 
	{tikz_figures/data/varying_pois_runtime_overhead.csv};
	\addplot[forget plot, draw=none,opacity=0.4,pattern=north east lines,fill=NavyBlue]
	fill between[of=top and bot];
	\addplot
	table [x=pois,y=Rome_avg,col sep=comma] 
	{tikz_figures/data/varying_pois_runtime_overhead.csv};
	\addplot[name path=top,draw=none,forget plot]
	table [name path=top,x=pois,y expr=\thisrow{Rome_avg}+\thisrow{Rome_std},col sep=comma] 
	{tikz_figures/data/varying_pois_runtime_overhead.csv};
	\addplot[name path=bot,draw=none,forget plot]
	table [name path=top,x=pois,y expr=\thisrow{Rome_avg}-\thisrow{Rome_std},col sep=comma] 
	{tikz_figures/data/varying_pois_runtime_overhead.csv};
	\addplot [forget plot, draw=none,opacity=0.4,pattern=north east lines,fill=orange]
	fill between[of=top and bot];
        \legend{Paris, Rome}
	\end{axis}
	\end{tikzpicture}
    \caption{Runtime.}
    \label{fig:scalability_POI_runtime}
    \end{subfigure}\hfill
\begin{subfigure}{0.33\textwidth}
    \centering
    \begin{tikzpicture}
	\begin{axis}[
	width=\textwidth,
	height=4.5cm,
	xmin=6,
	xmax=10,
	ymin=0,
	ymax=1.1,
	%
	%
	xlabel=Travel Hours,
	ylabel=\gpt Valid Plans,
	%
	mark options={scale=0.3},
	%
	%
	%
	]
	\addplot
	table [x=hours,y=Paris,col sep=comma] 
	{tikz_figures/data/varying_hours_invalidity.csv};
	\addplot
	table [x=hours,y=Rome,col sep=comma] 
	{tikz_figures/data/varying_hours_invalidity.csv};
        \legend{Paris, Rome}
	\end{axis}
	\end{tikzpicture}
    \caption{Plans validity.}
    \label{fig:scalability_hours_validity}
    \end{subfigure}\hfill
    \begin{subfigure}{0.33\textwidth}
    \centering
    \begin{tikzpicture}
	\begin{axis}[
	width=\textwidth,
	height=4.5cm,
	xmin=6,
	xmax=10,
	ymin=1,
	ymax=1.8,
 legend style={at={(axis cs:8,1.55)},anchor=south west},
	%
	%
	xlabel=Travel Hours,
	ylabel=\gpt Suboptimality,
	%
	mark options={scale=0.3},
	%
	%
	%
	]
	\addplot
	table [x=hours,y=Paris_avg,col sep=comma] 
	{tikz_figures/data/varying_hours_suboptimality.csv};
	\addplot[name path=top,draw=none,forget plot]
	table [name path=top,x=hours,y expr=\thisrow{Paris_avg}+\thisrow{Paris_std},col sep=comma] 
	{tikz_figures/data/varying_hours_suboptimality.csv};
	\addplot[name path=bot,draw=none,forget plot]
	table [name path=top,x=hours,y expr=\thisrow{Paris_avg}-\thisrow{Paris_std},col sep=comma] 
	{tikz_figures/data/varying_hours_suboptimality.csv};
	\addplot[forget plot, draw=none,opacity=0.4,pattern=north east lines,fill=NavyBlue]
	fill between[of=top and bot];
	\addplot
	table [x=hours,y=Rome_avg,col sep=comma] 
	{tikz_figures/data/varying_hours_suboptimality.csv};
	\addplot[name path=top,draw=none,forget plot]
	table [name path=top,x=hours,y expr=\thisrow{Rome_avg}+\thisrow{Rome_std},col sep=comma] 
	{tikz_figures/data/varying_hours_suboptimality.csv};
	\addplot[name path=bot,draw=none,forget plot]
	table [name path=top,x=hours,y expr=\thisrow{Rome_avg}-\thisrow{Rome_std},col sep=comma] 
	{tikz_figures/data/varying_hours_suboptimality.csv};
	\addplot [forget plot, draw=none,opacity=0.4,pattern=north east lines,fill=orange]
	fill between[of=top and bot];
        \legend{Paris, Rome}
	\end{axis}
	\end{tikzpicture}
    \caption{Plans suboptimality.}
    \label{fig:scalability_hours_suboptimality}
    \end{subfigure}\hfill
    \begin{subfigure}{0.33\textwidth}
    \centering
    \begin{tikzpicture}
	\begin{axis}[
	width=\textwidth,
	height=4.5cm,
	xmin=6,
	xmax=10,
	ymin=-200,
	ymax=100,
         legend style={at={(axis cs:6.5,-150)},anchor=south west},
	%
	%
	xlabel=Travel Hours,
	ylabel=\approach Extra Seconds,
	%
	mark options={scale=0.3},
	%
	%
	%
	]
	\addplot
	table [x=hours,y=Paris_avg,col sep=comma] 
	{tikz_figures/data/varying_hours_runtime_overhead.csv};
	\addplot[name path=top,draw=none,forget plot]
	table [name path=top,x=hours,y expr=\thisrow{Paris_avg}+\thisrow{Paris_std},col sep=comma] 
	{tikz_figures/data/varying_hours_runtime_overhead.csv};
	\addplot[name path=bot,draw=none,forget plot]
	table [name path=top,x=hours,y expr=\thisrow{Paris_avg}-\thisrow{Paris_std},col sep=comma] 
	{tikz_figures/data/varying_hours_runtime_overhead.csv};
	\addplot[forget plot, draw=none,opacity=0.4,pattern=north east lines,fill=NavyBlue]
	fill between[of=top and bot];
	\addplot
	table [x=hours,y=Rome_avg,col sep=comma] 
	{tikz_figures/data/varying_hours_runtime_overhead.csv};
	\addplot[name path=top,draw=none,forget plot]
	table [name path=top,x=hours,y expr=\thisrow{Rome_avg}+\thisrow{Rome_std},col sep=comma] 
	{tikz_figures/data/varying_hours_runtime_overhead.csv};
	\addplot[name path=bot,draw=none,forget plot]
	table [name path=top,x=hours,y expr=\thisrow{Rome_avg}-\thisrow{Rome_std},col sep=comma] 
	{tikz_figures/data/varying_hours_runtime_overhead.csv};
	\addplot [forget plot, draw=none,opacity=0.4,pattern=north east lines,fill=orange]
	fill between[of=top and bot];
        \legend{Paris, Rome}
	\end{axis}
	\end{tikzpicture}
    \caption{Runtime.}
    \label{fig:scalability_hours_runtime}
    \end{subfigure}\hfill
\caption{\gpt ratio of invalid plans (first column), \gpt suboptimality ratio (second column) and \approach extra seconds compared to \gpt (third column) as we increase the number of POIs (first row) and the travel hours $H$ (second row).}
\label{scalability}
\end{figure*}

Next, we evaluate how both approaches scale in terms of execution time and validity/quality of the returned travel plans as we increase the number of POIs and the  tourism hours.
We perform this analysis in two popular cities, Paris and Rome, and as before, we do $5$ runs for each $\langle \mbox{city}, \#\mbox{POIs}, H \rangle$.
We first fix the tourism hours to $8$ and increase the number of POIs from $8$ to $18$ in steps of $2$.
Then, we fix the number of POIs to $10$, and increase the tourism hours from $6$ to $10$.
The results of this evaluation are shown in Figure~\ref{scalability}.
As a reminder, all travel plans generated by \approach are guaranteed to be valid and optimal, so we will solely focus on \gpt's performance when referring to these two metrics.

As we can see in Figures~\ref{fig:scalability_POI_validity} and~\ref{fig:scalability_hours_validity}, \gpt is not able to consistently generate valid plans even in the simpler settings where it only needs to generate itineraries to visit $4$ POIs (Figure~\ref{fig:scalability_POI_validity}). While the $5$ plans to visit Paris are valid, none of the itineraries satisfy the hard constraints in the case of Rome.
The ratio of valid plans decreases even more as \gpt needs to consider more POIs, and we can observe the same behavior when we fix the number of POIs to $10$ and increase the travel hours (Figure~\ref{fig:scalability_hours_validity}).

Switching the focus to the quality of the returned plans (second column of Figure~\ref{scalability}), we can observe a clear decrease in the quality of the travel plans generated by \gpt as we increase the number of POIs to be considered, even when we remove the hard constraints checks.
\gpt is able to return the optimal solution in problems with only $4$ POIs, as the optimal travel plan in these cases simply involves visiting them all.
With $5$ and $6$ POIs, \gpt returns few plans with higher quality than \approach (suboptimality ratio $<1$).
In these cases, it is not properly reasoning about the hard constraints, and it is just fitting more POIs into the plan, at the price of generating plans that are not actually executable in practice.
However, the moment the problems start being oversubscribed, i.e., not all POIs can be visited within the given tourism hours, \gpt starts generating worse solutions that are up to $1.6$ times as bad as the optimal plan returned by \approach, which also offers the validity guarantees \gpt lacks.
The suboptimality trend is not that clear when we increase the travel hours (Figure~\ref{fig:scalability_hours_suboptimality}), as this allows \gpt to fit more POIs in a plan, thus increasing its quality.

Finally, the third column of Figure~\ref{scalability} shows the extra seconds required by \approach to generate a travel plan compared to \gpt.
As we can see in Figure~\ref{fig:scalability_POI_runtime}, both approaches running time is almost the same up to around $10$ POIs.
The resulting planning problems in these cases are small, and the planner can solve then in $<1$ second, so the main runtime driver is the interaction with the LLM.
As we increase the number of POIs, optimally solving the resulting tasks becomes more challenging, and \approach requires from few seconds to up to $800$ seconds more than \gpt to generate an optimal itinerary in Rome with $18$ POIs.
We do not observe this extra overhead when we fix the number of POIs and increase the travel hours (Figure~\ref{fig:scalability_hours_runtime}), as the planning tasks remain simple for the planner.
There was one anomalous data point for travel planning in Paris for 10 hours, in which the LLM approach took 285 seconds. This was due to network issues and delays, as the other 9 out of 10 tries for the same experimental setting took no more than 11 seconds.

\section{Discussion}
As mentioned in the introduction, we are using the utilities of POIs provided by \gpt as a proxy for user satisfaction when visiting each POI. But, utility based on tourist consensus (as retrieved by \gpt) can be replaced by user preferences. The user could state their preference in natural language, and \gpt can be used to find and rate places according to the user input. For example, if the user said they like art museums, then The Louvre would be highly rated. Other scoring methods maybe employed as well, and these new utilities can be input to the planner.

In our work, we limited ourselves to day-planning and discretized time in 15 minute segments. One can extend this to continuous time and multi-day planning at the expense of higher planning time costs. The value of more granular time travel plans is left for future work. Additionally one can build multi-day travel plans by simply chaining single day travel plans and excluding the POI visited in previous days. 

\section{Conclusions and Future Work}
In this paper, we presented a hybrid approach for travel planning that combines the strengths of LLMs and automated planners to generate travel plans that guarantee feasibility and maximize the satisfaction of user goals. We investigated the use of LLMs for oversubscription planning in the travel planning domain, considering real-world constraints. Experimental results across various travel scenarios demonstrate how we can greatly improve the performance of \gpt-based approaches in this domain by embedding them with a hybrid approach. 

This research opens up several exciting future directions. One avenue for exploration is integrating real-time information, such as traffic conditions and weather forecasts, into the planning process to further enhance the practicality and adaptability of the generated plans. Moreover, extending the hybrid approach to other domains beyond travel planning, such as event scheduling and resource allocation, could broaden the impact of this research. As the field of AI continues to advance, we believe that the synergistic combination of LLMs and planners will play a crucial role in enabling intelligent and user-centric decision-making systems.


\section{Disclaimer}
This paper was prepared for informational purposes by
the Artificial Intelligence Research group of JPMorgan Chase \& Co. and its affiliates (``JP Morgan''),
and is not a product of the Research Department of JP Morgan.
JP Morgan makes no representation and warranty whatsoever and disclaims all liability,
for the completeness, accuracy or reliability of the information contained herein.
This document is not intended as investment research or investment advice, or a recommendation,
offer or solicitation for the purchase or sale of any security, financial instrument, financial product or service,
or to be used in any way for evaluating the merits of participating in any transaction,
and shall not constitute a solicitation under any jurisdiction or to any person,
if such solicitation under such jurisdiction or to such person would be unlawful.

\bibliography{bibliography,daniel,general,travel-planning}

\end{document}


%

\begin{frontmatter}
\paperid{1549}
\title{Supplemental to \approach: Travel Planning with Guarantees by Combining Large Language Models and Automated Planners}
\author{Authors}


\end{frontmatter}




\section{ Prompt Engineering for Single Day Travel Plans}
In this section we present the prompts we tried. For each step we used the first prompt in the associated list. 
We employed different techniques learned to highlight information such as using two asterisk around words ($**<words>**$) to highlight points as training data for LLM includes markdown files in which the asterisks are used to make words bold. We also tried variations such as changing representation time from AM and PM to military time to get more valid travel plans from \gpt.

With each prompt, the particular format that we ask the output to be in, is to help us parse the output and build python data structures, which is subsequently fed into the UPF library for automated planning. The city name and other variable values used in the prompt are grounded examples; in the code, the city name and other variables are inserted for each experiment.

\subsection{Prompts for Generating POIs}
\begin{itemize}
    \item \textit{Give me a list of 10 tourist points of interest by their full name for the city of Paris. Present it as a comma separated list of places like placeA, place B, place C. No numbers.}    
    \item \textit{Give me a list of 10 tourist places for the city of Paris. Present it as a comma separated list of places like placeA, place B, place C. No numbers.}
    \item \textit{Give me a list of 10 tourist places for the city of Paris}
\end{itemize}
\subsection{Prompts for Obtaining POI Utility}
\begin{itemize}
    \item \textit{Given, this information, for the places mentioned, assign a number from 1 to 5 based on how popular they are for tourists. For example location one = 2 $\backslash{n}$ location two = 4 $\backslash{n}$ location three = 1 ... put each entry on a new line}
    \item \textit{"For the places mentioned, assign a number from 1 to 5 based on how popular they are for tourists. For example, location one = 2, location two = 4, location three = 1 ..."}
\end{itemize}
\subsection{Prompts for Obtaining Time to Spend in POI}

The prompt and output of the generated POI list is fed in as the prefix context, before the next prompt asking for utility is added. The prompts we tried to get utility values are below.
\begin{itemize}
    \item \textit{For the places and popularity mentioned, assign the amount of time one should spend at each of the locations.
        The amount of time should be in chunks of 15 minutes, and give the time in minutes not hours.For example
        location one = 15 minutes $\backslash{n}$ location two = 30 minutes $\backslash{n}$ location three = 75 minutes ... put each entry on a new line}
    \item \textit{For the places and popularity mentioned, assign the amount of time one should spend at each of the locations. The amount of time should be in chunks of 15 minutes, and give the time in minutes not hours. For example, location one = 15 minutes, location two = 30 minutes, location three = 75 minutes ... put each entry on a new line}
    
\end{itemize}


\subsection{Prompts for Generating Travel Plan in \gpt}

This was perhaps the most challenging step or prompt to adjust. We leveraged techniques like emphasizing words using $**$ to get better overall outcomes. 

\begin{itemize}
    \item \textit{Given the following information after the output format below, generate a travel plan for day in Paris
         between  8:00 and 16:00.
         The plan does ** NOT ** have to visit every place, only that the travel time starts at  8:00 and ends at 16:00, 
         and better places are visited based on the rating information given, and the duration of the visit conforms to the information below. $\backslash{n}$
         Ignore meal times, just focus on the places to visit$\backslash{n}$
         Assume we start at the location of the first place. The format of the plan should be like
         (visit place$\_$1 9:00 ) $\backslash{n}$ (drive place$\_$1 to place$\_$2 10:00 ) $\backslash{n}$ (visit place$\_$2 11:30 ) (drive place$\_$2 to place$\_$3 13:30 ) ...and so forth $\backslash{n}$ 
         for the time in each step, only specify the end time of the activity and assume the first one starts at 8:00 
         If a place has spaces in the name, replace with underscore$\backslash{n}$          
         DO NOT put numbers in the list ordering and do not write anything before or after the travel plan.$\backslash{n}$}
         
    \item \textit{Given the following information after the output format below, generate a travel plan for a day in Paris between 8:00 AM and 4:00 PM. The plan does NOT have to visit every place, only that the travel time is between 8:00 AM and 4:00 PM and better places are visited. Ignore meal times and exact time of day, just the order of places to visit.
    Assume we start at the location of the first place. The format of the plan should be like (visit place$\_$1 9:00) (drive place$\_$1 to place$\_$2 10:00) (visit place$\_$2 11:30) (drive place$\_$2 to place$\_$3 1:30) ...and so forth. For the time in each step, only specify the end time of the activity and assume the first one starts at 8:00. If a place has spaces in the name, replace with underscore. DO NOT put numbers in the list ordering and do not write anything before or after the travel plan}
   
\end{itemize}


\section {Full Example of Plan Generation}

In this section we present an end-to-end example of the travel planning process using \gpt and the \approach method.
We use the setting of a single-day travel plan for Madrid with 10 hours of total travel time. We ask \gpt to get a set of 10 points of interest along with travel times between them, visit times, and a rating for tourist value. We chose not to arbitrarily set ratings for places so as to align with information in \gpt; this is to make case-based planning easier for \gpt. 

\begin{itemize}

    \item Obtaining POI: 
    \item Obtaining ratings for POI: 
    \item Obtaining visit times for POI: 
    \item Determining travel times between POI: 
    
\end{itemize}

Thus far we have extracted the fundamental information needed to generate a travel plan. Next we query \gpt for it's travel plan and compare against the plan generated by an automated planner using the same information. 

The Query used for \gpt was:\textit{}



\subsection{Prompt and Context for \gpt planning}

The prompt and context together was input to \gpt for asking for a travel plan

The prompt was
\textit{Given the following information after the output format below, generate a travel plan for day in Paris between 8:00 and 14:00. The plan does ** NOT ** have to visit every place, only that the travel time starts at 8:00 and ends at 14:00and better places are visited based on the rating information given, and the duration of the visit conforms to the information below. 
Ignore meal times, just focus on the places to visit
Assume we start at the location of the first place. The format of the plan should be like
 (visit place$\_$1 9:00 ) $\backslash{n}$ (drive place$\_$1 to place$\_$2 10:00 ) $\backslash{n}$ (visit place$\_$2 11:30 ) (drive place$\_$2 to place$\_$3 13:30 ) ...and so forth $\backslash{n}$ 
 for the time in each step, only specify the end time of the activity and assume the first one starts at 8:00 
 If a place has spaces in the name, replace with underscore$\backslash{n}$          
 DO NOT put numbers in the list ordering and do not write anything before or after the travel plan.$\backslash{n}$}

The context provided to \gpt for planning is below, and this follows the prompt we gave for asking GPT to plan.

\textit{Place of interest are 
Eiffel Tower, Louvre Museum, Notre-Dame Cathedral, Sacré-Cœur Basilica, Palace of Versailles, Champs-Élysées, Montmartre, Sainte-Chapelle, Centre Pompidou, Musée d'Orsay
 The ratings of these place are 
Eiffel Tower = 5
Louvre Museum = 5
Notre-Dame Cathedral = 4
Sacré-Cœur Basilica = 4
Palace of Versailles = 5
Champs-Élysées = 4
Montmartre = 3
Sainte-Chapelle = 3
Centre Pompidou = 2
Musée d'Orsay = 4
 The time needed to vist these place are 
Eiffel Tower = 120 minutes
Louvre Museum = 180 minutes
Notre-Dame Cathedral = 60 minutes
Sacré-Cœur Basilica = 60 minutes
Palace of Versailles = 240 minutes
Champs-Élysées = 90 minutes
Montmartre = 75 minutes
Sainte-Chapelle = 45 minutes
Centre Pompidou = 90 minutes
Musée d'Orsay = 120 minutes
 The travel time to go between each of these place are as follows,and assume the travel time is in multiples of 15 minutes rounding up, so the minimum is 15 minutes.
Travel time from eiffel tower, Paris to louvre museum, Paris is 8 mins
Travel time from eiffel tower, Paris to notre$\_$dame cathedral, Paris is 13 mins
Travel time from eiffel tower, Paris to sacr$\_$cur basilica, Paris is 14 mins
Travel time from eiffel tower, Paris to palace of versailles, Paris is 35 mins
Travel time from eiffel tower, Paris to champs$\_$lyses, Paris is 14 mins
Travel time from eiffel tower, Paris to montmartre, Paris is 23 mins
Travel time from eiffel tower, Paris to sainte$\_$chapelle, Paris is 16 mins
Travel time from eiffel tower, Paris to centre pompidou, Paris is 18 mins
Travel time from eiffel tower, Paris to muse dorsay, Paris is 9 mins
Travel time from louvre museum, Paris to eiffel tower, Paris is 11 mins
Travel time from louvre museum, Paris to notre$\_$dame cathedral, Paris is 6 mins
Travel time from louvre museum, Paris to sacr$\_$cur basilica, Paris is 8 mins
Travel time from louvre museum, Paris to palace of versailles, Paris is 45 mins
Travel time from louvre museum, Paris to champs$\_$lyses, Paris is 8 mins
Travel time from louvre museum, Paris to montmartre, Paris is 23 mins
Travel time from louvre museum, Paris to sainte$\_$chapelle, Paris is 8 mins
Travel time from louvre museum, Paris to centre pompidou, Paris is 11 mins
Travel time from louvre museum, Paris to muse dorsay, Paris is 4 mins
Travel time from notre$\_$dame cathedral, Paris to eiffel tower, Paris is 15 mins
Travel time from notre$\_$dame cathedral, Paris to louvre museum, Paris is 7 mins
Travel time from notre$\_$dame cathedral, Paris to sacr$\_$cur basilica, Paris is 7 mins
Travel time from notre$\_$dame cathedral, Paris to palace of versailles, Paris is 49 mins
Travel time from notre$\_$dame cathedral, Paris to champs$\_$lyses, Paris is 7 mins
Travel time from notre$\_$dame cathedral, Paris to montmartre, Paris is 27 mins
Travel time from notre$\_$dame cathedral, Paris to sainte$\_$chapelle, Paris is 2 mins
Travel time from notre$\_$dame cathedral, Paris to centre pompidou, Paris is 10 mins
Travel time from notre$\_$dame cathedral, Paris to muse dorsay, Paris is 8 mins
Travel time from sacr$\_$cur basilica, Paris to eiffel tower, Paris is 17 mins
Travel time from sacr$\_$cur basilica, Paris to louvre museum, Paris is 10 mins
Travel time from sacr$\_$cur basilica, Paris to notre$\_$dame cathedral, Paris is 3 mins
Travel time from sacr$\_$cur basilica, Paris to palace of versailles, Paris is 46 mins
Travel time from sacr$\_$cur basilica, Paris to champs$\_$lyses, Paris is 1 mins
Travel time from sacr$\_$cur basilica, Paris to montmartre, Paris is 25 mins
Travel time from sacr$\_$cur basilica, Paris to sainte$\_$chapelle, Paris is 5 mins
Travel time from sacr$\_$cur basilica, Paris to centre pompidou, Paris is 8 mins
Travel time from sacr$\_$cur basilica, Paris to muse dorsay, Paris is 11 mins
Travel time from palace of versailles, Paris to eiffel tower, Paris is 32 mins
Travel time from palace of versailles, Paris to louvre museum, Paris is 37 mins
Travel time from palace of versailles, Paris to notre$\_$dame cathedral, Paris is 42 mins
Travel time from palace of versailles, Paris to sacr$\_$cur basilica, Paris is 43 mins
Travel time from palace of versailles, Paris to champs$\_$lyses, Paris is 43 mins
Travel time from palace of versailles, Paris to montmartre, Paris is 46 mins
Travel time from palace of versailles, Paris to sainte$\_$chapelle, Paris is 44 mins
Travel time from palace of versailles, Paris to centre pompidou, Paris is 46 mins
Travel time from palace of versailles, Paris to muse dorsay, Paris is 37 mins
Travel time from champs$\_$lyses, Paris to eiffel tower, Paris is 17 mins
Travel time from champs$\_$lyses, Paris to louvre museum, Paris is 10 mins
Travel time from champs$\_$lyses, Paris to notre$\_$dame cathedral, Paris is 3 mins
Travel time from champs$\_$lyses, Paris to sacr$\_$cur basilica, Paris is 1 mins
Travel time from champs$\_$lyses, Paris to palace of versailles, Paris is 46 mins
Travel time from champs$\_$lyses, Paris to montmartre, Paris is 25 mins
Travel time from champs$\_$lyses, Paris to sainte$\_$chapelle, Paris is 5 mins
Travel time from champs$\_$lyses, Paris to centre pompidou, Paris is 8 mins
Travel time from champs$\_$lyses, Paris to muse dorsay, Paris is 11 mins
Travel time from montmartre, Paris to eiffel tower, Paris is 24 mins
Travel time from montmartre, Paris to louvre museum, Paris is 20 mins
Travel time from montmartre, Paris to notre$\_$dame cathedral, Paris is 27 mins
Travel time from montmartre, Paris to sacr$\_$cur basilica, Paris is 25 mins
Travel time from montmartre, Paris to palace of versailles, Paris is 48 mins
Travel time from montmartre, Paris to champs$\_$lyses, Paris is 25 mins
Travel time from montmartre, Paris to sainte$\_$chapelle, Paris is 29 mins
Travel time from montmartre, Paris to centre pompidou, Paris is 23 mins
Travel time from montmartre, Paris to muse dorsay, Paris is 22 mins
Travel time from sainte$\_$chapelle, Paris to eiffel tower, Paris is 14 mins
Travel time from sainte$\_$chapelle, Paris to louvre museum, Paris is 7 mins
Travel time from sainte$\_$chapelle, Paris to notre$\_$dame cathedral, Paris is 3 mins
Travel time from sainte$\_$chapelle, Paris to sacr$\_$cur basilica, Paris is 5 mins
Travel time from sainte$\_$chapelle, Paris to palace of versailles, Paris is 47 mins
Travel time from sainte$\_$chapelle, Paris to champs$\_$lyses, Paris is 5 mins
Travel time from sainte$\_$chapelle, Paris to montmartre, Paris is 25 mins
Travel time from sainte$\_$chapelle, Paris to centre pompidou, Paris is 8 mins
Travel time from sainte$\_$chapelle, Paris to muse dorsay, Paris is 8 mins
Travel time from centre pompidou, Paris to eiffel tower, Paris is 18 mins
Travel time from centre pompidou, Paris to louvre museum, Paris is 11 mins
Travel time from centre pompidou, Paris to notre$\_$dame cathedral, Paris is 3 mins
Travel time from centre pompidou, Paris to sacr$\_$cur basilica, Paris is 2 mins
Travel time from centre pompidou, Paris to palace of versailles, Paris is 47 mins
Travel time from centre pompidou, Paris to champs$\_$lyses, Paris is 2 mins
Travel time from centre pompidou, Paris to montmartre, Paris is 26 mins
Travel time from centre pompidou, Paris to sainte$\_$chapelle, Paris is 6 mins
Travel time from centre pompidou, Paris to muse dorsay, Paris is 11 mins
Travel time from muse dorsay, Paris to eiffel tower, Paris is 11 mins
Travel time from muse dorsay, Paris to louvre museum, Paris is 5 mins
Travel time from muse dorsay, Paris to notre$\_$dame cathedral, Paris is 9 mins
Travel time from muse dorsay, Paris to sacr$\_$cur basilica, Paris is 10 mins
Travel time from muse dorsay, Paris to palace of versailles, Paris is 45 mins
Travel time from muse dorsay, Paris to champs$\_$lyses, Paris is 10 mins
Travel time from muse dorsay, Paris to montmartre, Paris is 23 mins
Travel time from muse dorsay, Paris to sainte$\_$chapelle, Paris is 10 mins
Travel time from muse dorsay, Paris to centre pompidou, Paris is 14 mins
}
\bibliography{bibliography,daniel,general,travel-planning}